\documentclass{vgtc}                          %
\graphicspath{{figures/}{pictures/}{images/}{./}} %

\usepackage{times}                     %
\usepackage{tabu}                      %
\usepackage{booktabs}                  %
\usepackage{lipsum}                    %
\usepackage{mwe}                       %

\usepackage{mathptmx}                  %

\onlineid{0}

\vgtccategory{Research}

\vgtcinsertpkg

\usepackage{comment}
\usepackage{xcolor}
\usepackage{xspace}
\usepackage{enumitem}
\usepackage{wrapfig}
\usepackage{caption}
\usepackage{subcaption}

\definecolor{myACMBlue}{cmyk}{1,0.1,0,0.1}
\definecolor{myACMYellow}{cmyk}{0,0.16,1,0}
\definecolor{myACMOrange}{cmyk}{0,0.42,1,0.01}
\definecolor{myACMRed}{cmyk}{0,0.90,0.86,0}
\definecolor{myACMLightBlue}{cmyk}{0.49,0.01,0,0}
\definecolor{myACMGreen}{cmyk}{0.20,0,1,0.19}
\definecolor{myACMPurple}{cmyk}{0.55,1,0,0.15}
\definecolor{myACMDarkBlue}{cmyk}{1,0.58,0,0.21}
\definecolor{niceblue}{RGB}{56, 116, 203}

\hypersetup{colorlinks,
      linkcolor=niceblue,
      citecolor=niceblue,
      urlcolor=niceblue,
      filecolor=niceblue}

\definecolor{red}{HTML}{df2c14}
\definecolor{pink}{HTML}{f86f92}
\definecolor{hotpink}{HTML}{fe036a}
\definecolor{green}{HTML}{75975e}
\definecolor{skyblue}{HTML}{00cbfe}
\definecolor{blue}{HTML}{1167b1}
\definecolor{cornflowerblue}{HTML}{797ef6}
\definecolor{navy}{HTML}{194569}
\definecolor{violet}{HTML}{710193}
\definecolor{purple}{HTML}{a32cc4}
\definecolor{lilac}{HTML}{b65fcf}
\definecolor{addedblue}{HTML}{247ce0}
\definecolor{deletedred}{HTML}{eb4034}
\definecolor{brightyellow}{HTML}{ffb300}

\definecolor{textgreen}{HTML}{4d9221}
\definecolor{imagepink}{HTML}{c51b7d}
\definecolor{origorange}{HTML}{d6604d}
\definecolor{compblue}{HTML}{4393c3}

\newcommand{\de}[0]{Diffusion Explainer\xspace}
\newcommand{\sd}[0]{Stable Diffusion\xspace}
\newcommand{\rcv}[0]{Refinement Comparison View\xspace}
\newcommand{\trajectory}[0]{Evolution Trajectory\xspace}

\usepackage{ulem}

\renewcommand{\sectionautorefname}{\S\kern-1pt}
\renewcommand{\subsectionautorefname}{\S\kern-1pt}
\renewcommand{\subsubsectionautorefname}{\S\kern-1pt}

\renewcommand\footnotemark{}

\title{\de: Visual Explanation for Text-to-image Stable Diffusion}

\newcommand{\authorgap}{\hspace{10pt}}
\author{
Seongmin Lee\textsuperscript{\textrm *}
\thanks{\textsuperscript{\textrm *}Georgia Tech. 
\{\href{mailto:seongmin@gatech.edu}{seongmin}$\mid$%
\href{mailto:bhoov@gatech.edu}{bhoov}$\mid$%
\href{mailto:jayw@gatech.edu}{jayw}$\mid$%
\href{mailto:speng65@gatech.edu}{speng65}$\mid$%
\href{mailto:apwright@gatech.edu}{apwright}$\mid$%
\href{mailto:kevin.li@gatech.edu}{kevin.li}$\mid$
\href{mailto:haekyu@gatech.edu}{haekyu}$\mid$%
\href{mailto:alexanderyang@gatech.edu}{alexanderyang}$\mid$%
\href{mailto:polo@gatech.edu}{polo}\}@gatech.edu} \authorgap
Benjamin Hoover \textsuperscript{\textrm *,\textdagger}
\thanks{\textsuperscript{\textrm \textdagger}IBM Research. 
\href{mailto:hendrik.strobelt@ibm.com}{hendrik.strobelt@ibm.com}
} \authorgap
Hendrik Strobelt\textsuperscript{\textrm \textdagger} \authorgap 
Zijie J. Wang\textsuperscript{\textrm *}\authorgap 
ShengYun Peng\textsuperscript{\textrm *}\authorgap \\
Austin Wright\textsuperscript{\textrm *}\authorgap
Kevin Li\textsuperscript{\textrm *}\authorgap 
Haekyu Park\textsuperscript{\textrm *} \authorgap 
Haoyang Yang\textsuperscript{\textrm *}\authorgap 
Duen Horng (Polo) Chau\textsuperscript{\textrm *}
}

\teaser{
  \centering
  \includegraphics[width=\textwidth]{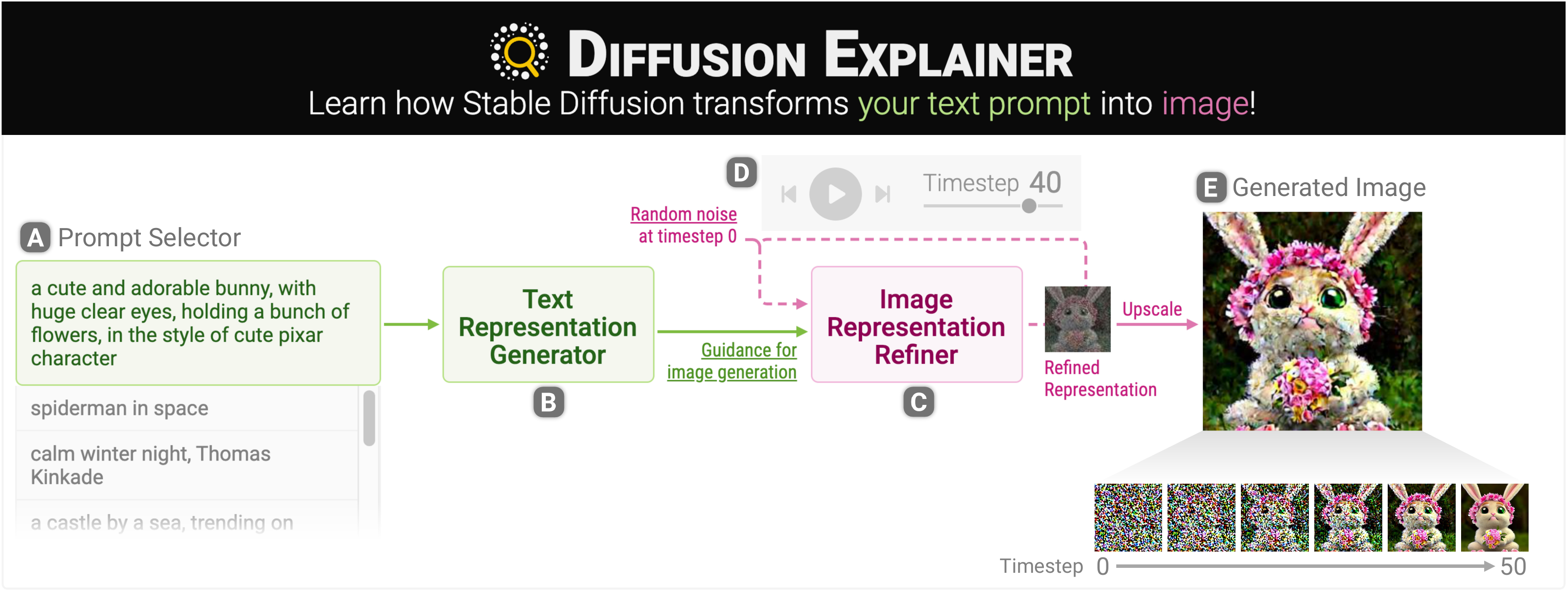}
  \vspace{-15pt}
  \caption{
  With \de, users can visually examine how
  \textbf{(A)} \textit{text prompt}, e.g., \textit{``a cute and adorable bunny... pixar character''}, is encoded by 
  \textbf{(B)} the \textit{Text Representation Generator} into vectors to guide 
  \textbf{(C)} the \textit{Image Representation Refiner} to iteratively refine the vector representation of the image being generated.
  \textbf{(D)} The \textit{Timestep Controller} enables users to review the incremental improvements in image quality and adherence to the prompt over timesteps. 
  \textbf{(E)} The final image representation is upscaled to a high-resolution image.
  Diffusion Explainer tightly integrates 
  a visual overview of \sd's complex components with detailed explanations of their underlying operations,
  enabling users to fluidly transition between multiple levels of abstraction through animations and interactive elements (\autoref{fig:text}, \autoref{fig:image}).
  }
  \label{fig:architecture}
}

\abstract{
  Diffusion-based generative models' impressive ability to create convincing images has garnered global attention.
  However, their complex structures and operations often pose challenges for non-experts to grasp.
  We present \de, the first interactive visualization tool that explains how \sd transforms text prompts into images.
  \de tightly integrates a visual overview of Stable Diffusion’s complex structure with explanations of the underlying operations. 
  By comparing image generation of prompt variants, users can discover the impact of keyword changes on image generation.
  A 56-participant user study
  demonstrates that \de offers substantial learning benefits to non-experts.
  Our tool has been used by over 10,300 users from 124 countries at \url{https://poloclub.github.io/diffusion-explainer/}.
} %

\keywords{}

\begin{document}

\maketitle

\section{Introduction}
\label{sec:intro}
Diffusion-based generative models~\cite{rombach2022high,stability2022stable,openai2022dalle}
like Stable Diffusion~\cite{stability2022stable} and DALL-E~\cite{openai2022dalle} 
have captured global attention for their impressive image creation abilities, from AI developers, designers, to policymakers.
However, the popularity and progress of generative AI models have sparked social concerns~\cite{sung2022lensa,brusseau2022acceleration,choudhary2022stable,dixit2023meet}, such as accusations of artistic style theft by developers of AI image generators~\cite{choudhary2022stable,dixit2023meet}.
Policymakers are also discussing ways to combat malicious data generation and revise copyright policies~\cite{eshoo2022,engler2023early,ryanmosley2023an,2023copyright}.
There is an urgent need for individuals from many different fields to understand how generative AI models function and communicate effectively with AI researchers and developers~\cite{dixit2023meet,hendrix2023generative}. 

\smallskip
\noindent 
\textbf{Key challenges in designing learning tools for \sd.}
At the high level, \sd iteratively refines noise into a vector representation of a high-resolution image, guided by a text prompt. 
Internally, the prompt is tokenized and encoded into vector representations by \textit{CLIP Text Encoder}~\cite{radford2021learning}.
With the text representations' guidance,
\sd improves the image quality and adherence to the prompt by progressively denoising the image's vector representation using the \textit{UNet} neural network~\cite{ronneberger2015u}.
The final image representation is upscaled to a high-resolution image~\cite{kingma2013auto}.
The crux of learning about \sd stems from the
complex interplay between multiple subcomponents, their intricate operations, and the iterative nature of refinements,
which are challenging even for  experts to grasp~\cite{platen2022testing}. 
While some articles~\cite{alammar2022the} and videos~\cite{howard2023from,alammar2023ai} explain \sd,
they often assume notable knowledge of machine learning and focus on %
mathematical details.

In this work, we contribute:
\begin{itemize}[topsep=2pt, itemsep=0mm, parsep=3pt, leftmargin=10pt]
    \item \textbf{\de, the first interactive visualization tool designed for non-experts}
    to explain how \sd transforms a text prompt into a high-resolution image, overcoming key design challenges in developing interactive learning tools for \sd (\autoref{fig:architecture}).
    It tightly integrates a visual overview of \sd's complex structure with detailed explanations of their underlying operations via animations and interactive elements (\autoref{fig:text}, \autoref{fig:image}; \autoref{sec:architecture}).
    It also provides a new way to visualize the impact of keyword changes on the complex image generation process
    by comparing image generation of prompt variants (\autoref{fig:rcv}; \autoref{sec:rcv}).
    \item \textbf{Reflection and design lessons derived from human evaluation with 56 non-experts} showcase
     \de's substantial advantages in 
    explaining \sd to non-experts,  compared to the common blog post approach.  
    The great majority preferred \de, 
    rating it easier to understand and more effective for improving their learning.
    We distill key design lessons for creating visualizations to educate non-experts on modern AI techniques (\autoref{sec:userstudy}).
    \item \textbf{A web-based implementation}
    that broadens the public's education access to modern generative AI techniques without requiring any installation, advanced hardware, or coding skills.
    \de runs locally in users' browsers, enabling a large number of concurrent users to learn directly on their devices (\autoref{subsec:410:overview}).
    Available at \url{https://poloclub.github.io/diffusion-explainer/}, 
    \de is open-source\footnote{\url{https://github.com/poloclub/diffusion-explainer}}.
    Having been used by over 10,300 users from 124 countries, \de is 
    making 
    strides in democratizing AI education.
\end{itemize}

\section{Related Works}
\label{sec:related}
\textbf{Interactive Visualizations for Explaining Deep Learning.}
Early tools like
ConvNetJS MNIST demo~\cite{karpathy2023convnetjs} and Tensorflow Playground~\cite{smilkov2017direct} enabled users to experiment with simple models and datasets directly in their browsers.
To explain more advanced techniques, 
researchers have developed interactive articles~\cite{goh2017momentum,carter2017using,madsen2019visualizing,agnihotri2020exploring,sanchez2021gentle,olah2023colah},
but they often assume prior machine learning knowledge.
To address the needs of non-experts,
interactive visualization tools
such as CNN Explainer~\cite{wangCNNExplainerLearning2020},
GAN Lab~\cite{kahng2018ganlab}, 
and
Adversarial-Playground~\cite{norton2017adversarial} were developed.
Inspired by their success, we develop \de as a web-based interactive visualization to broaden %
education access to \sd.

\smallskip
\noindent
\textbf{Explanations for Stable Diffusion.}
Online articles that explain \sd
often assume machine learning expertise, using jargons and equations that can be daunting for non-experts~\cite{weng2021diffusion,alammar2022the,hosni2022getting,andrew2023how},
while the articles for beginners~\cite{bogaard2022an,andrew2023absolute} mainly address deployment and prompt engineering.
Additionally,
they focus on either high-level structures~\cite{bogaard2022an,andrew2023absolute}
or low-level operations~\cite{hosni2022getting,andrew2023how}, %
overlooking the need for a comprehensive understanding~\cite{kahng2018ganlab,wangCNNExplainerLearning2020}.
Google Colab tutorials~\cite{patil2022stable,whitaker2022grokking} %
require %
coding skills, posing challenges in learning.
\de enables easy experimentation without coding,
offering clear explanations for \sd's architecture and operations through interactive elements.

\section{Design Goals}
\label{sec:300:designgoals}

By reviewing literature~\cite{rombach2022high,platen2022testing}, 
we
established four design goals: %

\begin{enumerate}[topsep=1pt, itemsep=0mm, parsep=1pt, leftmargin=19pt, label=\textbf{G\arabic*.}, ref=G\arabic*]
    \item \label{goal:summary} \textbf{Visual summary of \sd.}
    \sd involves multiple complex model components~\cite{rombach2022high,platen2022testing} and cyclic refinement from noise to the vector representation of a high-resolution image.
    \de provides
    an overview of the model architecture and cyclic data flow to help users quickly understand its overall structure~\cite{kahng2017activis} (\autoref{sec:architecture}). 
    \item \label{goal:interactive} \textbf{Interactive interface tightly integrating different abstraction levels.} 
    \sd's image generation process is hard to comprehend
    due to a complex interplay between multiple subcomponents and their intricate underlying operations~\cite{platen2022testing,radford2021learning,rombach2022high}.
    To effectively explain the low-level operations and %
    conceptually connect them with a high-level overview,
    we bridge multiple abstraction levels through fluid animations and interactive elements~\cite{wangCNNExplainerLearning2020,kahng2018ganlab} (\autoref{sec:architecture}).
    \item \label{goal:refine} \textbf{Visualizing how keyword changes in text prompts affect image generation.} 
    Modifying a few keywords from prompts can unexpectedly lead to dramatic changes in generated images~\cite{liu2022design} (e.g., repeating ``\textit{very}'' multiple times~\cite{oppenlaender2022taxonomy}), making it important for users to gain awareness and understanding of such impact on the generated images~\cite{feng2023promptmagician}.
    We visualize the refinement process for two text prompts that differ only in a few keywords to compare how image representations evolve differently when guided by each prompt (\autoref{sec:rcv}).
    \item \label{goal:webbased} \textbf{Broadening access via web-based deployment.}
    As many individuals from various fields are interested in understanding generative AI~\cite{dixit2023meet,engler2023early,ryanmosley2023an}, we develop \de to run locally on users' devices without requiring any installation, specialized hardware, or coding.
    To offer real-time interactive learning experiences~\cite{wangCNNExplainerLearning2020}, we pre-compute intensive processes on predetermined prompts (\autoref{subsec:410:overview}),
    instead of generating images for user-provided prompts using the nascent WebGPU technology~\cite{ninomiya2024webgpu}, which is limited in both browser support and speed\footnote{At the time of writing, WebGPU is only supported in Chrome browser and requires restricted settings to run \sd~\cite{machine2024web}. Generating a single image takes at least a few minutes, with even longer times on CPUs. Training \sd, which is even more computationally demanding, is therefore infeasible.}.
    We open-source our code for easy extension to new prompts and hyperparameter settings.
\end{enumerate}

\section{System Design and Implementation}
\label{sec:design}
\subsection{Overview}
\label{subsec:410:overview}
\de is a web-based interactive visualization that explains how \sd generates a high-resolution image from a text prompt.
It incorporates an animation of random noise gradually refined and a \textit{Timestep Controller} (\autoref{fig:architecture}D) that enables users to visit each refinement timestep.
From the \textit{Prompt Selector} (\autoref{fig:architecture}A), users select one out of the 13 prompts that follow a template~\cite{smith2022traveler} and include popular keywords (e.g., \textit{detailed}, \textit{trending on artstation})
identified from literature~\cite{oppenlaender2022taxonomy,andrew2023stable,pavlichenko2022best}.
\de consists of two views: 
\textit{Architecture View} (\autoref{sec:architecture}) tightly integrates a visual overview of \sd's architecture (\ref{goal:summary}) and the underlying operations via interactive elements and animation (\ref{goal:interactive}), and 
\textit{\rcv} (\autoref{sec:rcv}) compares image generation of two related prompts to uncover the impact of prompt keywords on image generation (\ref{goal:refine}).
Below the visualization, we provide text explanations for more details. \de is implemented using a standard web technology stack (HTML, CSS, JavaScript) and the D3.js~\cite{bostock2011d3} library (\ref{goal:webbased}).

\subsection{Architecture View} 
\label{sec:architecture}
The Architecture View shows a visual overview (\ref{goal:summary}, \autoref{fig:architecture}) of how \textit{Text Representation Generator} (\autoref{fig:architecture}A) converts a text prompt into vector representations that guide \textit{Image Representation Refiner} (\autoref{fig:architecture}B) to incrementally refine noise into the vector representation of a high-resolution image. Users can click the generators to expand them into more details about their underlying operations.

\begin{figure}[t]
    \centering
    \includegraphics[width=0.93\columnwidth]{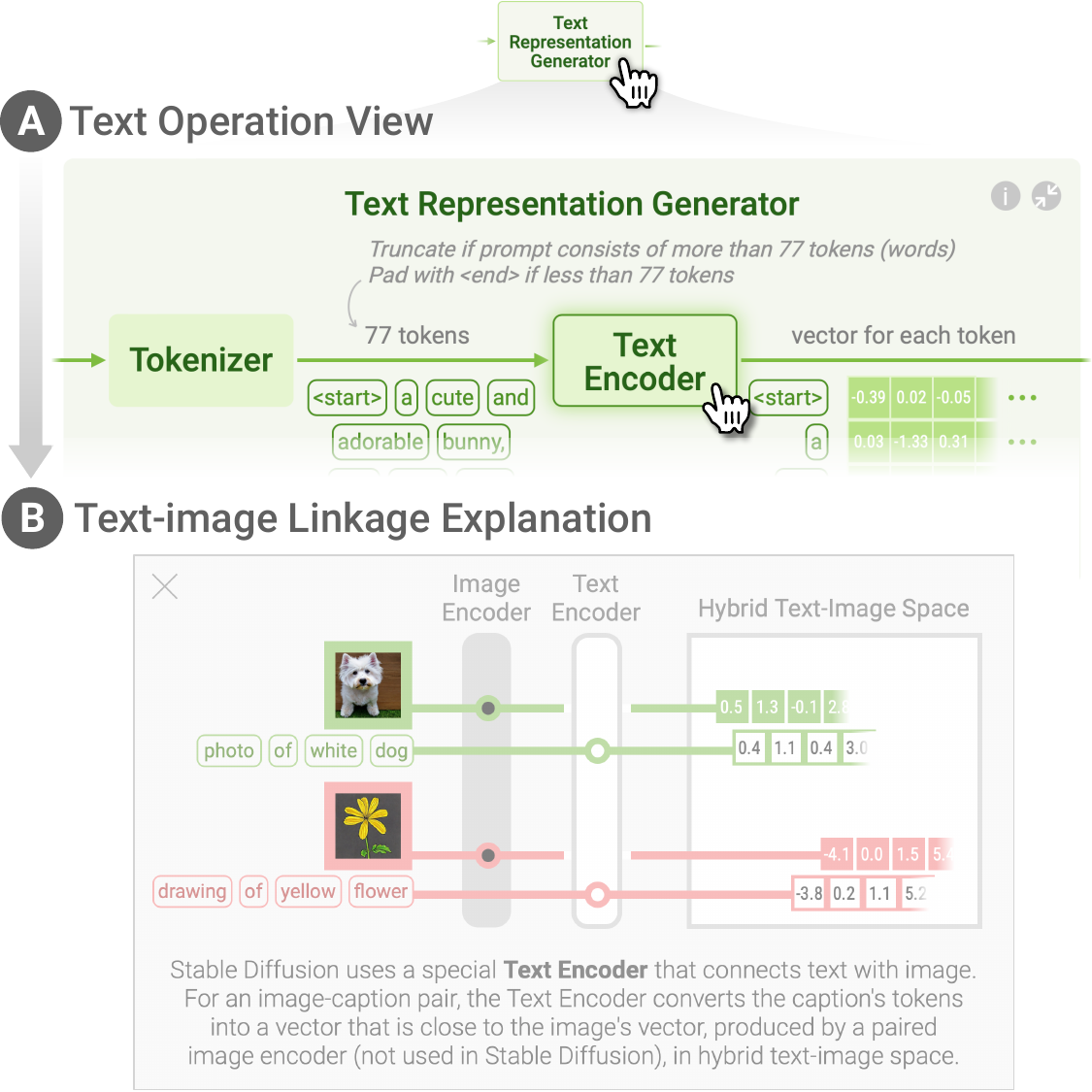}
    \vspace{-2pt}
    \caption{
    To learn how \sd converts a text prompt into vector representations, users click the \textit{Text Representation Generator}, which smoothly expands to
    \textbf{(A)} the \textit{Text Operation View}, which explains how the prompt is split into tokens and encoded into vector representations.
    \textbf{(B)} The \textit{Text-image Linkage Explanation}
    demonstrates how \sd bridges text and image, 
    enabling text representations to guide the image generation process.
    }
    \vspace{-15pt}
    \label{fig:text}
\end{figure}

\begin{figure}[t]
    \centering
    \includegraphics[width=0.93\columnwidth]{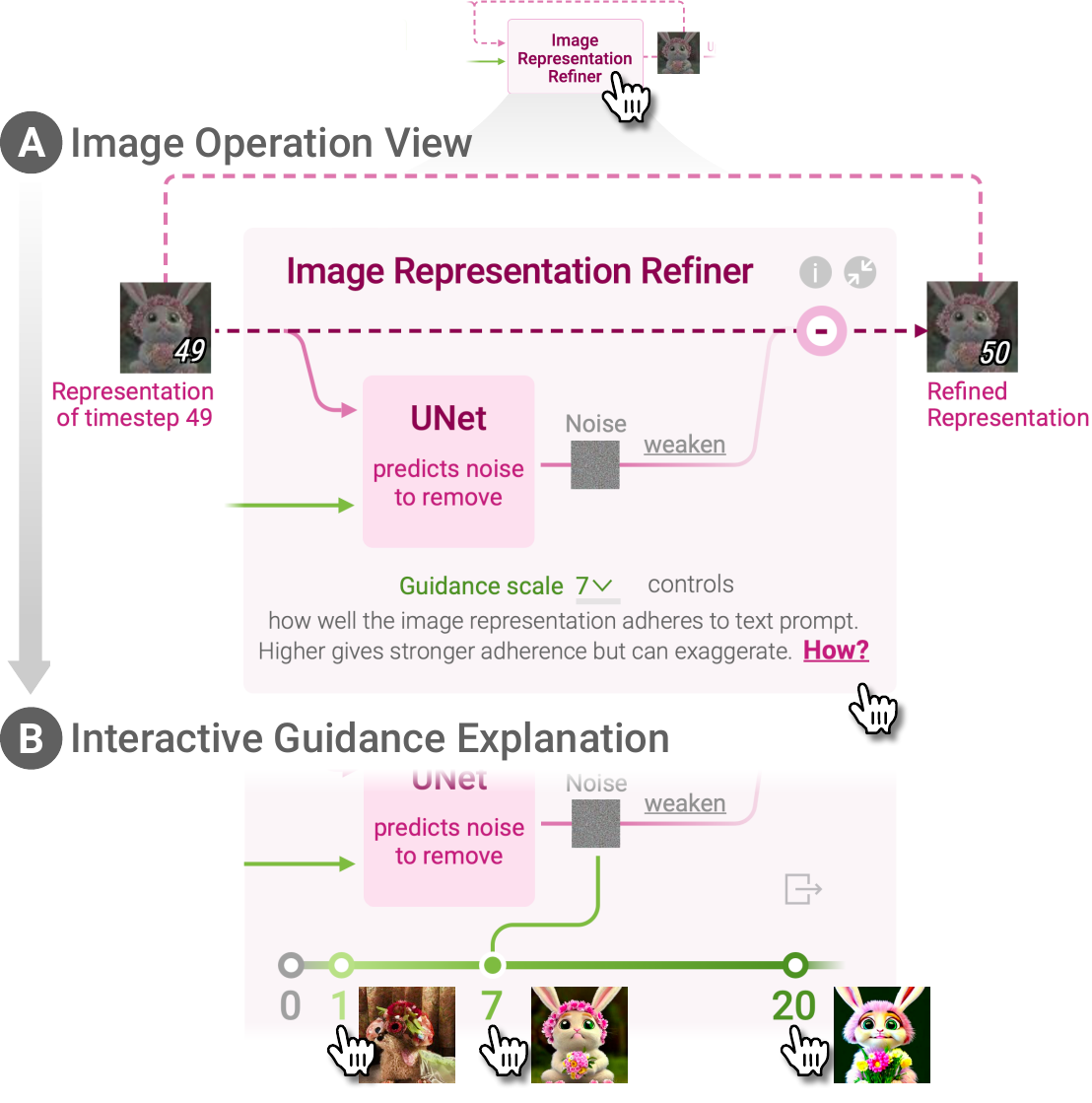}
    \vspace{-2pt}
    \caption{
    Users learn how \sd refines noise into a high-resolution image's vector representation aligned with the text prompt by clicking the \textit{Image Representation Refiner} to smoothly expand to 
    \textbf{(A)} the \textit{Image Operation View}
    that demonstrates how noise is predicted and removed from the image representation.
    \textbf{(B)} The \textit{Interactive Guidance Explanation} allows users to interactively experiment with different guidance scale values (0, 1, 7, 20) 
    to better understand how higher values lead to stronger adherence.}
    \vspace{-15pt}
    \label{fig:image}
\end{figure}

The \textit{Text Representation Generator} converts a text prompt into vector representations. Users can click and expand it to the \textit{Text Operation View} (\ref{goal:interactive}, \autoref{fig:text}A) to learn that
the prompt is split into tokens, and then the tokens are encoded into vector representations using Text Encoder.
Clicking on the Text Encoder displays the \textit{Text-image Linkage Explanation} (\ref{goal:interactive}; \autoref{fig:text}B), 
which illustrates that 
CLIP~\cite{radford2021learning} text encoder generates text representations with image-related information, which is crucial for guiding image generation.
The \textit{Image Representation Refiner} incrementally refines random noise into the vector representation of an image that adheres to the text prompt.
The image representation of each refinement step is visualized by (1) decoding into a small image using linear operations~\cite{turner2022decoding} and 
(2) upscaling to \sd's output resolution.
Clicking the Image Representation Refiner expands it to the \textit{Image Operation View} (\ref{goal:interactive}; \autoref{fig:image}A),
which explains that the refinement consists of noise prediction and removal.
The guidance scale hyperparameter, which controls the image's adherence strength to the text prompt, is described at the bottom, and further explained in the \textit{Interactive Guidance Explanation} (\ref{goal:interactive}; \autoref{fig:image}B).
It provides a slider to experiment with different guidance scale values to better understand how higher values lead to stronger adherence.

\begin{figure*}
    \centering
    \includegraphics[width=0.8\linewidth]{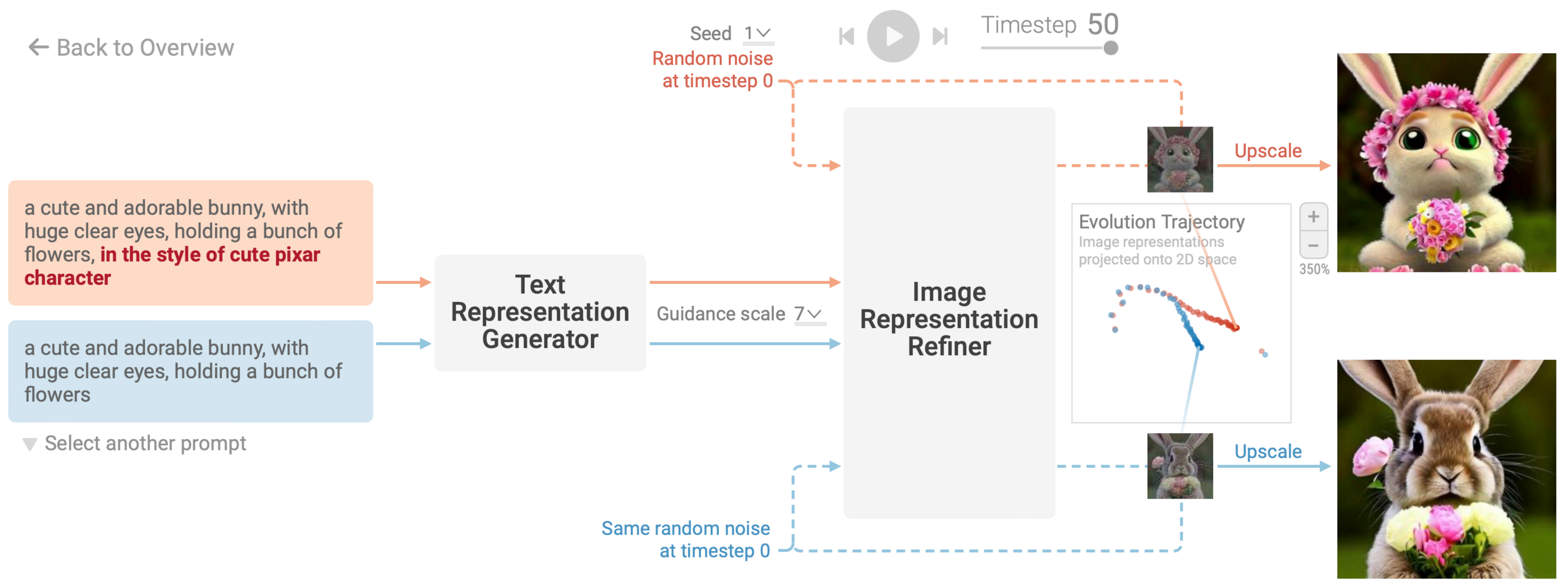}
    \vspace{-5pt}
    \caption{
    The \textit{\rcv} enables users to discover the impacts of prompts on image generation by comparing how image representations evolve differently over refinement timesteps, using UMAP, when guided by two related text prompts. 
    Adding \textit{``pixar''} phrase changes the generated bunny's style to be more cartoony and vibrant in colors and textures while preserving its pose.
    }
    \label{fig:rcv}
    \vspace{-8pt}
\end{figure*}

\subsection{\rcv}
\label{sec:rcv}
The \textit{\rcv} demonstrates how modifying a few keywords in prompt can significantly change the generated images
(\ref{goal:refine}; \autoref{fig:rcv}).
Each prompt in \de is paired with a prompt that differs only in a few keywords; for example, an original prompt ``\textit{a cute and adorable bunny... \textbf{pixar character}}'' is paired with the prompt variant ``\textit{a cute and adorable bunny...}" (keywords highlighted in \textbf{bold}).
We smoothly animate the transitions between the \rcv and the Architecture View, %
preserving the location of
the original prompt and the model components, %
while fading in the prompt variant from underneath the original prompt.
We visualize the \textit{\trajectory} of image representations for the paired prompts
to reveal
how prompt keywords affect the evolution of image representations %
(\ref{goal:refine}).
Each point on the \trajectory corresponds to an image representation at each timestep,
and 
their progression over timesteps is animated.
We compute 2-dimensional representations of the image representations across all timesteps, text prompts, guidance scales, and random seeds 
using UMAP\footnote{For UMAP hyperparameters, we conducted extensive testing over a wide range of  \texttt{n\_neighbors} (5 to 30), \texttt{min\_dist} (0.1 to 0.99), and random seed (0, 1, 2), and did not observe a significant impact on the local structure. Hence, we employed a fixed configuration of \texttt{n\_neighbors} of 15, \texttt{min\_dist} of 0.99, 
and random seed of 0.} \cite{mcinnes2018umap}.

\section{Human Evaluation}
\label{sec:userstudy}
To evaluate the effectiveness of \de, %
we conducted a user study.
We followed a within-subjects design~\cite{seltman2012experimental} to compare \de with other publicly accessible explanations. %
\subsection{Procedure}
\label{sec:procedure}
We recruited participants %
from Prolific\footnote{\url{https://www.prolific.com}}, an online recruiting platform. 
After participants signed a consent document, we conducted a background survey.
To include only non-experts interested in image generative AI, we asked participants to indicate their interest in image generative AI and to self-report their knowledge of AI and image generative AI on a scale from 1 to 5. %
We qualified only those who expressed interest and self-identified as having little knowledge of AI and generative AI (rated as \textit{1: Don't know what it is}, \textit{2: Heard of it only}, or \textit{3: occasional users of AI-powered tools}).

Each participant then used \de and a top-ranked blog post\footnote{\url{http://jalammar.github.io/illustrated-stable-diffusion/}}  on Google, to learn about \sd.
We excluded other sources (e.g., videos, Google Colab tutorials) as they were promotional, not free, or required advanced hardware and coding skills.
To counterbalance the order effect, half of the participants began with \de, while the other half used it after the blog post.
To verify tool engagement, participants answered three simple quiz questions after using each tool\footnote{Quizzes are easy for participants who used the tools, e.g., ``Select all guidance scale values supported by \de''}. 
We considered only responses from participants who answered at least two questions correctly for both tools.
After using the two tools,
participants rated them by answering 5-point Likert-scale questions.

The study lasted approximately 40 minutes per participant, with each compensated \$10.00.
Fifty-six participants passed the knowledge and engagement screening.
On average, their self-evaluated knowledge level in AI was 2.79, and in image generative AI was 2.41.
Participants came from diverse fields, including art, education, public administration, health care, real estate, finance, retail, construction, and manufacturing.

\subsection{Results and Design Lessons}
\label{sec:results}
\begin{wrapfigure}{R}{0.50\columnwidth}
   \centering
   \includegraphics[width=0.50\columnwidth]{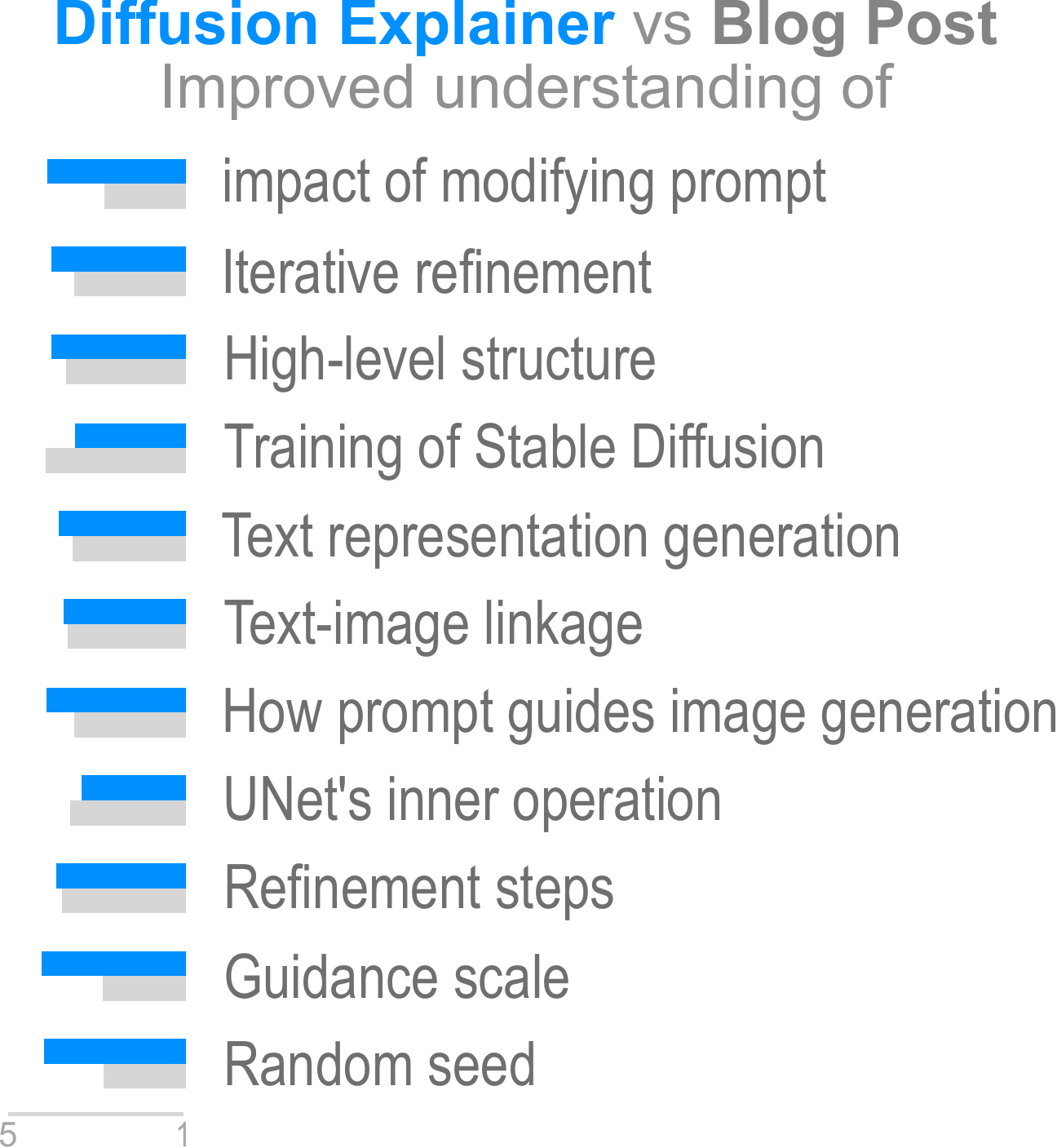}
   \vspace{-15pt}
   \caption{\de more effective than blog post for learning most \sd concepts.}
   \vspace{3pt}
   \label{fig:understanding}
\end{wrapfigure}

\vspace{1pt}
\setlength{\columnsep}{10pt}%
\setlength{\intextsep}{0.6pt}%
\begin{wrapfigure}{R}{0.45\columnwidth}
   \centering
   \includegraphics[width=0.45\columnwidth]{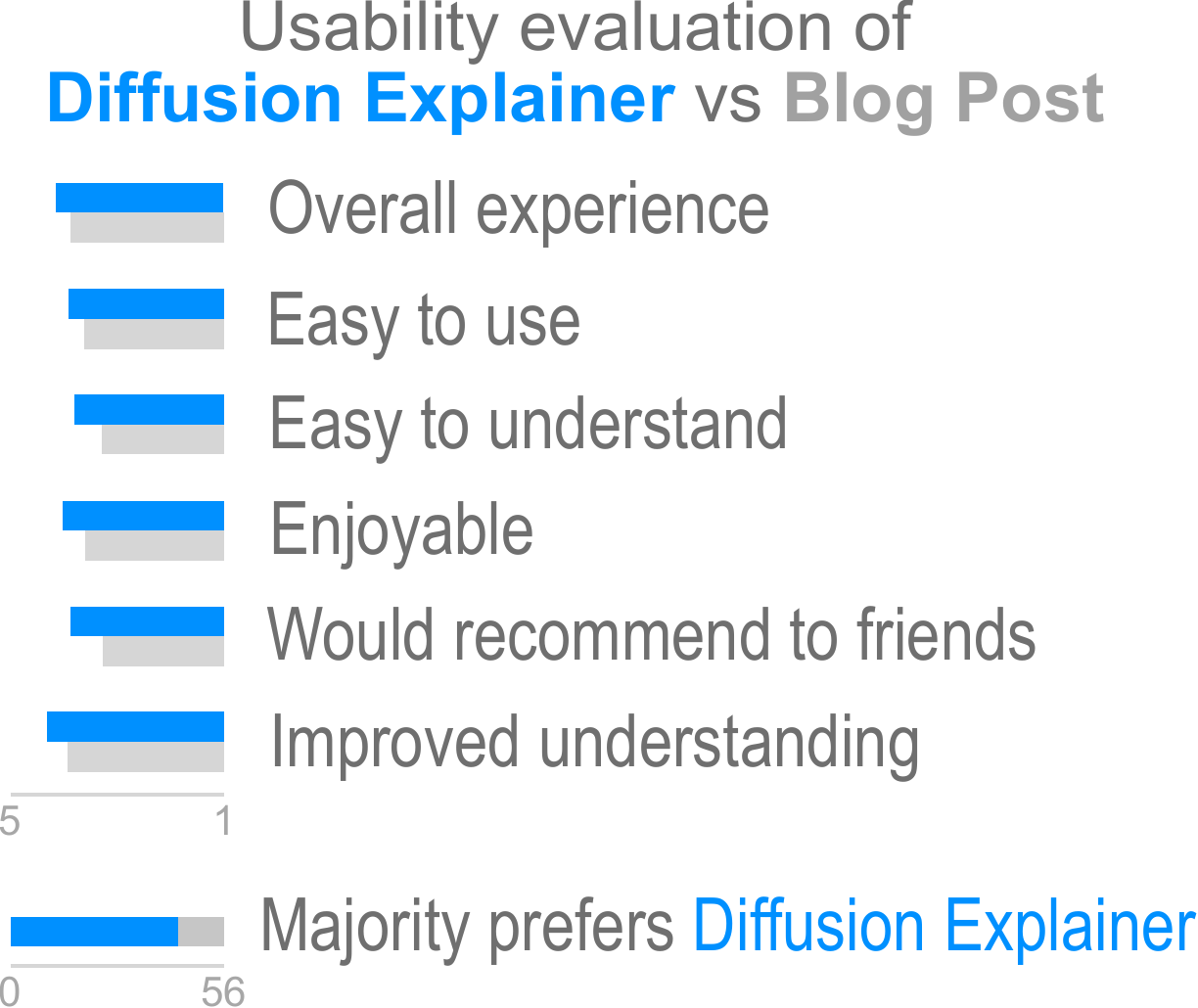}
   \vspace{-15pt}
   \caption{
   \de more usable than blog post.
   }
   \vspace{1pt}
   \label{fig:usability}
\end{wrapfigure}
\autoref{fig:usability} summarizes the 
participants' responses comparing the usability of the two tools. Overall, the great majority preferred \de over the blog post (44 out of 56 participants). They found \de more enjoyable, more helpful for improving their understanding, and easier to understand, even without expertise in AI. \\
\medskip
\noindent
\setlength{\columnsep}{7pt}%
\setlength{\intextsep}{0.6pt}%
\begin{wrapfigure}{R}{0.45\columnwidth}
   \centering
   \includegraphics[width=0.45\columnwidth]{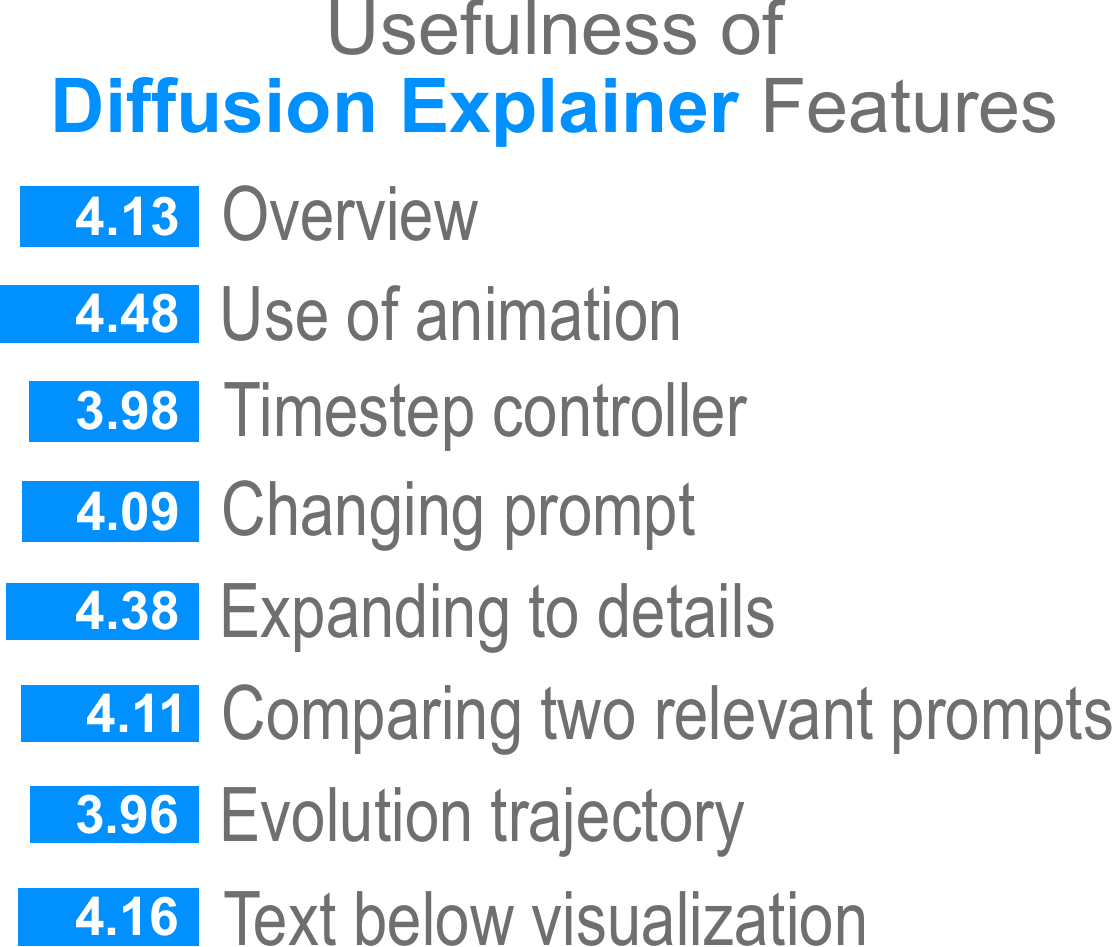}
   \vspace{-14pt}
   \caption{All features  were rated highly.}
   \vspace{1pt}
   \label{fig:feature}
\end{wrapfigure}
\textbf{Non-experts easily used \de, improving  understanding of \sd.}
We are excited to observe the high ratings given to all of \de's features  (\autoref{fig:feature}), showcasing its substantial educational advantages over the blog post (\autoref{fig:usability}, \autoref{fig:understanding}).
For example, when explaining how text prompts guide image generation, \de received an average score of 4.18, notably surpassing the blog post's 3.55 (\autoref{fig:understanding}).

\medskip
\noindent
\setlength{\columnsep}{7pt}%
\setlength{\intextsep}{0.6pt}%
\textbf{Thoughtful design  crucial for bridging  abstraction levels (\ref{goal:summary}, \ref{goal:interactive}).}
Our study confirms the key benefit of interactive design over a static post that can overwhelm users by presenting all information at once.
Participants appreciated \de's high-level overview (4.13) and animated expansions into details (4.48).
Our decision not to delve into advanced concepts, such as UNet's architecture and the training of \sd, as explained in depth in the blog post (\autoref{fig:understanding}), 
maintains \de's accessibility to non-experts.
One participant noted,  ``\textit{layers of the UNet noise predictor and jargons made [the blog post] too overwhelming and technical.}'' %
This underscores the importance of carefully selecting details to include in visualizations to balance the amount of information with the users' expertise.

\smallskip
\noindent
\textbf{Interactive visualization offers unique learning benefits for understanding hyperparameters (\ref{goal:interactive}).}
Participants found \de substantially more effective than the blog post
in explaining the guidance scale (4.29 vs. 2.91) and random seed (4.25 vs. 2.88)  (\autoref{fig:understanding}).
They appreciated the ability to rapidly experiment with \de. %
One participant noted, ``\textit{[it] is like a cool game changing settings to get good-looking pictures for each prompt, helping me learn about making images from text}.''
Also, participants could grasp the intricate interaction between prompts and hyperparameters that cannot be easily replicated by a static blog post 
Another participant mentioned, ``\textit{The guidance scale for the best image quality depends on the prompt. I should decide the value carefully for each prompt.}''

\smallskip
\noindent
\textbf{Comparing prompt variants helps non-experts understand impact of prompt keywords (\ref{goal:refine}).}
As a first tool designed to help non-experts learn about how slight changes in a prompt can significantly influence image generation, we are thrilled to learn that \de's \rcv was favorably rated. In particular, prompt comparison received an average score of 4.11, and the \trajectory powered by  UMAP~\cite{mcinnes2018umap}, a technique often considered for advanced analysis, was also well-rated (3.96).
The effectiveness is further demonstrated by \de's substantially higher rating (4.16 vs. 2.86) for improving understanding of the impact of prompt modification on image generation (\autoref{fig:understanding}).
A participant remarked,
``\textit{I like how Diffusion Explainer lets me see how slight changes in prompts affect the style and quality of pictures. It's cool to compare two different prompts and see how images and representations develop over time.}''

\section{Conclusion}
\label{sec:conclusion}
We have introduced \de, the first interactive web-based visualization tool that explains how \sd generates high-resolution images from text prompts. 
Our tool seamlessly integrates a visual overview with detailed explanations of the underlying operations
through animations and interactive elements.
Its innovative design uncovers the impacts of prompt keywords on image generation.
A user study with 56 non-experts demonstrates the superiority of \de over a popular blog post.
We hope our research  inspires further development of visualizations to enhance people's understanding of modern AI technologies.

\bibliographystyle{abbrv-doi}

\bibliography{references}
\end{document}